\documentclass[11pt,twocolumn,letterpaper]{article}

\usepackage{cvpr}
\usepackage{times}
\usepackage{epsfig}
\usepackage{graphicx}
\usepackage{amsmath}
\usepackage{amssymb}
\usepackage{array}
\newcolumntype{L}[1]{>{\raggedright\let\newline\\\arraybackslash\hspace{0pt}}m{#1}}

\usepackage[breaklinks=true,bookmarks=false]{hyperref}

\cvprfinalcopy 

\def\cvprPaperID{****} 

\begin{document}

\title{Hotel Recommendation System}

\author{
Aditi A. Mavalankar\\
A53200443 \\
{\tt\small amavalan@ucsd.edu}
\and
Ajitesh Gupta\\
A53220177 \\
{\tt\small ajgupta@ucsd.edu}
\and 
Chetan Gandotra\\
A53210397 \\
{\tt\small cgandotr@ucsd.edu}
\and
Rishabh Misra\\
A53205530\\
{\tt\small r1misra@ucsd.edu}
}

\maketitle

\begin{abstract}
   One of the first things to do while planning a trip is to book a good place to stay. Booking a hotel online can be an overwhelming task with thousands of hotels to choose from, for every destination. Motivated by the importance of these situations, we decided to work on the task of recommending hotels to users. We used Expedia's hotel recommendation dataset, which has a variety of features that helped us achieve a deep understanding of the process that makes a user choose certain hotels over others. The aim of this hotel recommendation task is to predict and recommend five hotel clusters to a user that he/she is more likely to book given hundred distinct clusters. 
\end{abstract}

\section{Introduction} \label{section:1}
Everyone likes their products to be personalized and behave the way they want them to. Given a user, recommender systems aim to model and predict the preference of a product. We study the Expedia online hotel booking system \footnote{https://www.expedia.com/Hotels} to recommend hotels to users based on their preferences. The dataset was made available by Expedia as a Kaggle challenge \footnote{https://www.kaggle.com/c/expedia-hotel-recommendations/}. Expedia wants to take the proverbial rabbit hole out of hotel search by providing personalized hotel recommendations to their users. Currently, Expedia uses search parameters to adjust their hotel recommendations, but there aren't enough customer specific data to personalize them for each user. 

In this competition, Expedia is challenging Kagglers to contextualize customer data and predict the likelihood a user will stay at 100 different hotel groups. The goal is to provide not just one recommendation, but to rank the predictions and return the top five most likely hotel clusters for each user's particular search query in the test set. We use multiple models and techniques to arrive at our best solution. This includes (1) An ensemble of four different models (random forests, SGD classifier, XG Boost and Naive Bayes), (2) XG Boost that is preceded by completion of distance matrix - an important feature in the dataset which is currently incomplete, (3) Data leakage solution which takes advantage of the fact that there is a potential leak in the data provided to us, and (4) a mixture of the methods in (1) and (3).

\section{Dataset} \label{section:2}
We have used the \href{https://www.kaggle.com/c/expedia-hotel-recommendations/data}{Expedia Hotel Recommendation dataset} from Kaggle. The dataset, which had been collected in the 2013-2014 time-frame, consists of a variety of features that could provide us great insights into the process user go through for choosing hotels. The training set consists of 37,670,293 entries and the test set contains 2,528,243 entries. Apart from this, the dataset also provide some latent features for each of the destinations recorded in the train and test sets. The data is anonymized and almost all the fields are in numeric format. The goal is to predict 5 hotel clusters where a user is more likely to stay, out of a total of 100 hotel clusters. The problem has been modeled as a ranked multi-class classification task. Missing data, ranking requirement, and the curse of dimensionality are the main challenges posed by this dataset.

\subsection{Features}
The features have been provided to us in two forms - (I) geographical, temporal and search features in one set, and (II) latent features about destinations extracted from hotel reviews text. The location and temporal information about users and hotel clusters has been provided in terms of region, country, continent, check-in/check-out dates etc. On the other hand, the 149 latent features have been extracted from hotel reviews on subjects relevant to hotel services such as cleanliness, proximity to attractions, etc. The description of type (I) features is provided in table \ref{table:1}.

\begin{table}
	\caption{Features used for Training and Testing}
    \vspace{1em}
	\begin{tabular}{| L{3.7cm} | L{3.8cm} |}
		\hline
		\bf{Feature} & \bf{Description} \\ 
		\hline
 		date\_time & Timestamp \\
        \hline
		site\_name & ID of Expedia point of sale \\
        \hline
		posa\_continent & ID of site's continent \\
        \hline
 		user\_location\_country & ID of customer's country \\
        \hline
 		user\_location\_region & ID of customer's region \\
        \hline
 		user\_location\_city & ID of customer's city \\
        \hline
 		orig\_destination\_distance & Physical distance between a hotel and a customer \\
        \hline
 		user\_id & ID of user \\
        \hline
 		is\_mobile & 1 for mobile device, 0 otherwise \\
        \hline
 		is\_package & 1 if booking/click was part of package, 0 otherwise \\
        \hline
 		channel & ID of a marketing channel \\
        \hline
 		srch\_ci & Check-in date \\
        \hline
 		srch\_co & Check-out date \\
        \hline
 		srch\_adults\_cnt & Number of adults \\
        \hline
        srch\_children\_cnt & Number of children \\
        \hline
        srch\_rm\_cnt & Number of rooms \\
        \hline
        srch\_destination\_id & ID of the destination \\
        \hline
        srch\_destination\_type\_id & Type of destination \\
        \hline
        hotel\_continent & Hotel continent \\
        \hline
        hotel\_country & Hotel country \\
        \hline
        hotel\_market & Hotel market \\
        \hline
        is\_booking & 1 if a booking, 0 if a click \\
        \hline
        cnt & Number of similar events in the context of the same user sessiont \\
        \hline
        hotel\_cluster & ID of hotel cluster \\
        \hline
 \end{tabular}
\label{table:1}
\end{table}

\subsection{Data Preprocessing and Exploratory Analysis} \label{section:2.2}

Our first step was to clean and pre-process the data and perform exploratory analysis to get some interesting insights into the process of choosing a hotel. 

The first thing we observed was that there were many users who have only searched for hotels and did not make any reservation. Moreover, the test data had only the users who made a reservation. Thus, we pruned the dataset by removing all the users who did not make any booking as these entries do not provide any indication of which hotel clusters those users prefer and this could possibly have interfered with making predictions.

From the remaining entries, we identified the searches by each user belonging to a specific type of destination. This gave us some useful information about which hotel cluster was finally chosen over other hotel clusters explored by the user. One important observation to note is that few users might be travel agents and could explore multiple type of destinations at the same time. This could also be true for few users who are planning multiple vacations at the same time. That is why we considered the preferences of the users separately for each destination type he/she explored. Also, after a booking was made, subsequent searches by the user were treated separately. We describe aforementioned approach with an example below.

\begin{table}[!ht]
\caption{Subset of training set} 
\centering 
\begin{tabular}{c c c c} 
\hline\hline 
User ID & Hotel Cluster & Destination Type & Booking \\ [0.5ex] 
\hline 
1 & 3 & 1 & 0 \\ 
1 & 56 & 1 & 0 \\
1 & 35 & 6 & 0 \\
1 & 23 & 1 & 1 \\
1 & 25 & 6 & 1 \\ [1ex] 
\hline 
\end{tabular}
\label{table:2}
\end{table}

In table \ref{table:2}, the searches made by user 1 are shown. There are 2 types of destinations: 1 and 6. Based on the type of destination, we identify the hotel clusters that were rejected/selected. For destination type 1, cluster 23 was selected, and clusters 3 and 56 were rejected. Similarly, for destination type 6, cluster 25 was selected and cluster 35 was rejected. Now, we keep a track of the rejected clusters and create a vector of dimension $1 \times 100$ where each index corresponds to a hotel cluster. The values at the indices of the rejected clusters is set to -1 and at the index of the selected cluster is set to 1. Other entries in the vector are set to 0. We add this vector as features in the our dataset. Now, since entries 1, 2, and 3 give no extra information we would remove them and augment the last two entries with corresponding newly generated vectors.

\begin{figure}[h!]
\begin{center}
\includegraphics[width=\linewidth]{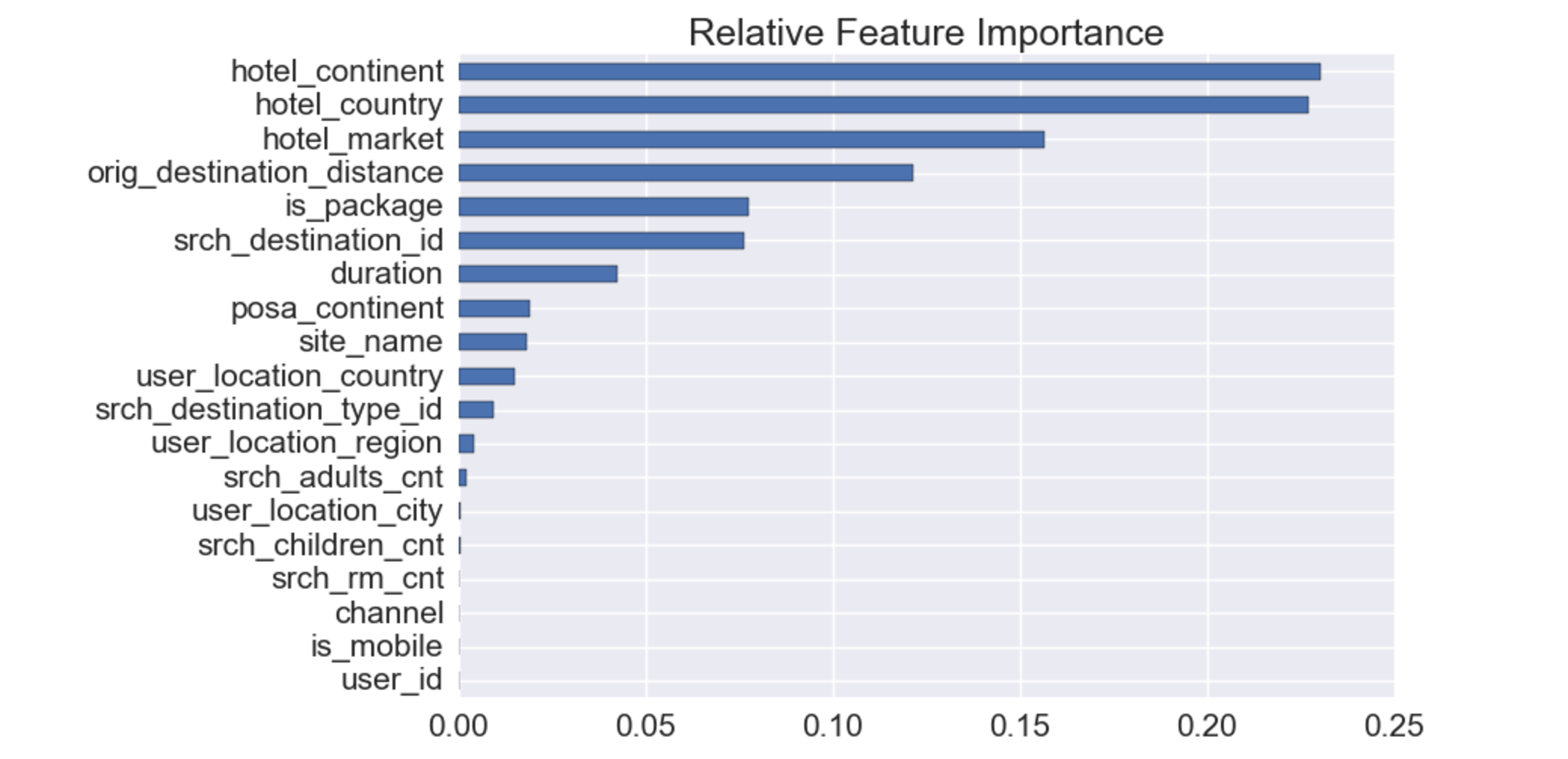}
\end{center}
   \caption{Relative Feature Importance}
\label{fig:feature_importance}
\end{figure}

Furthermore, we observed that the value of $orig\_destination\_distance$ attribute was missing in many entries in train and test sets. Since, this attribute is very important while recommending a hotel, as the distance between the hotel location and user location can give us a much better idea of the hotel cluster he/she may pick (reaffirmed by Figure \ref{fig:feature_importance}), we used matrix factorization technique to fill in the missing values in the dataset. We describe our approach in more detail in Section \ref{section:4.9}.

\begin{figure}[b]
\begin{center}
\includegraphics[width=\linewidth]{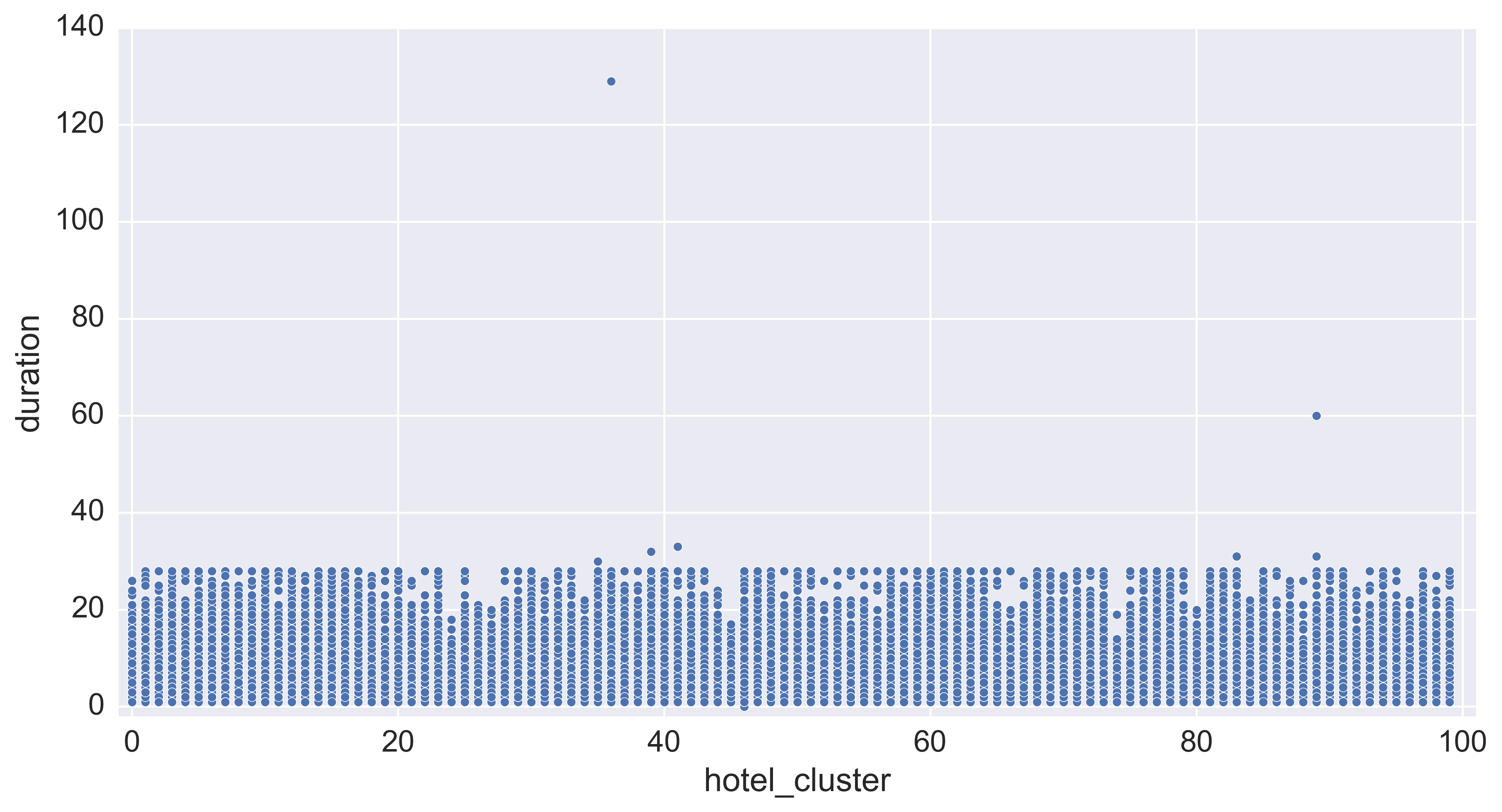}
\end{center}
   \caption{Duration of stay vs hotel cluster}
\label{fig:duration}
\end{figure}

Also, we use the check-in and check-out dates to find the duration of the stay for each of the entries in the training set. From the Figure \ref{fig:duration}, we see that all the clusters seem equally likely when the duration is short, but as the duration increases, certain hotel clusters are preferred over the others. This seems to be a good feature to identify the hotel cluster user would choose. This observation is reaffirmed in Figure \ref{fig:feature_importance}.

Furthermore, as per the details provided by Expedia on the competition page, hotels tend to change their cluster seasonally. To capture this variation, we also included one-hot representation of the month from which user is seeking to start his/her stay.

\begin{figure}[t]
\begin{center}
\includegraphics[width=\linewidth]{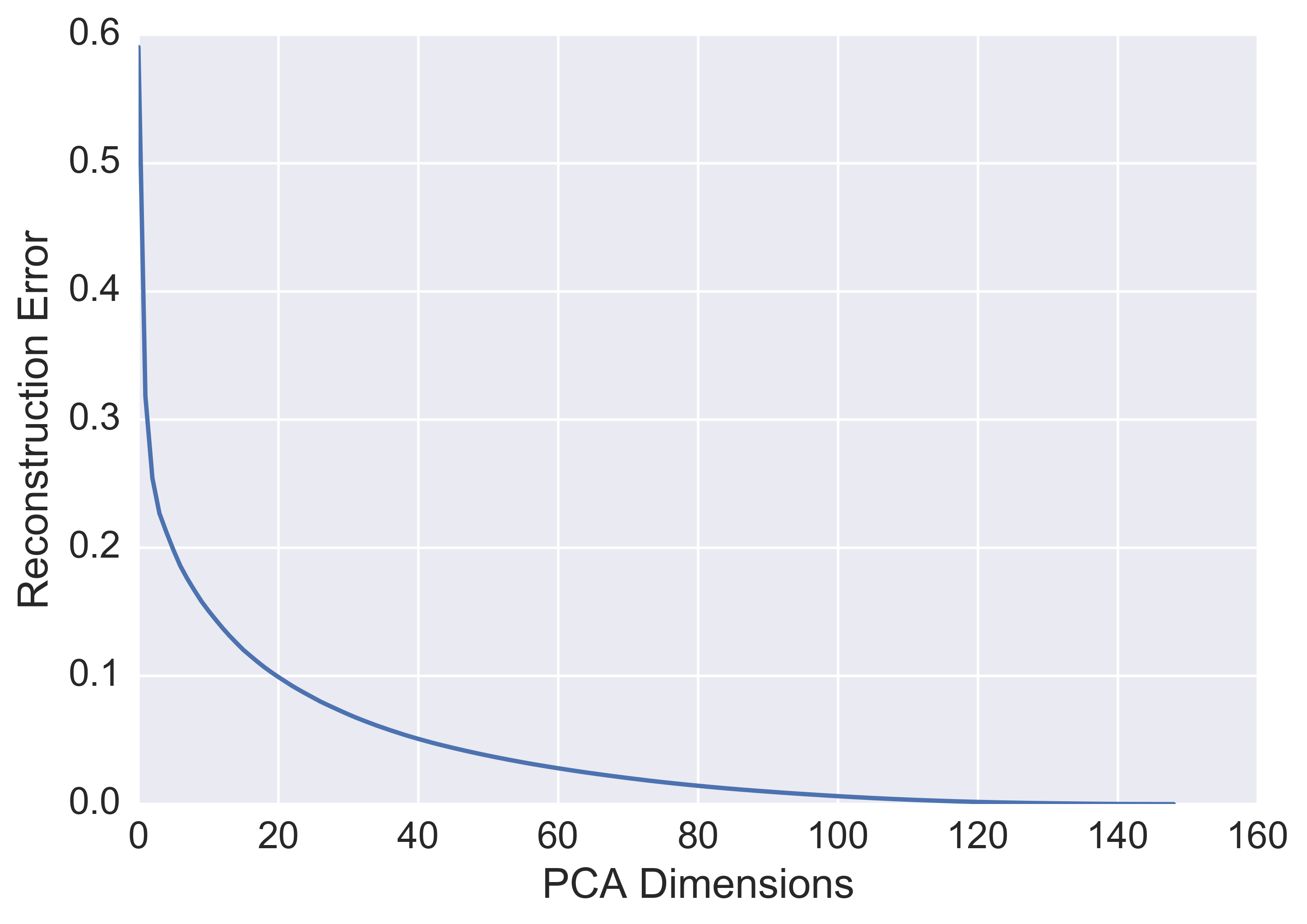}
\end{center}
   \caption{Reconstruction Error vs Latent dimensions}
\label{fig:reconstruction_error}
\end{figure}

Finally, we make use of the latent features of the destinations provided in the dataset. However, since we have 149 latent features for each destination, we decided to apply PCA to extract the most relevant dimensions. As Figure \ref{fig:reconstruction_error} suggests, we should select the number of component as 20 as further increasing the dimension does not reduce the reconstruction error much. We add these latent features along with our already prepared features.

Having prepared the extra set of features which we thought would be a good indicative of the user's choice of the hotel cluster, we perform the ablation experiment to gauge the importance of resulting features. We visualize the relative importance of features in Figure \ref{fig:feature_importance}. From the figure, we observe that geographical location and the distance between user and the hotel (which we calculated using distance matrix completion method) are the most important features. 

Next, we visualize the correlation matrix between the features of the training set in Figure \ref{fig:cor} and observe following things:
\begin{figure}[t]
\begin{center}
\includegraphics[width=\linewidth]{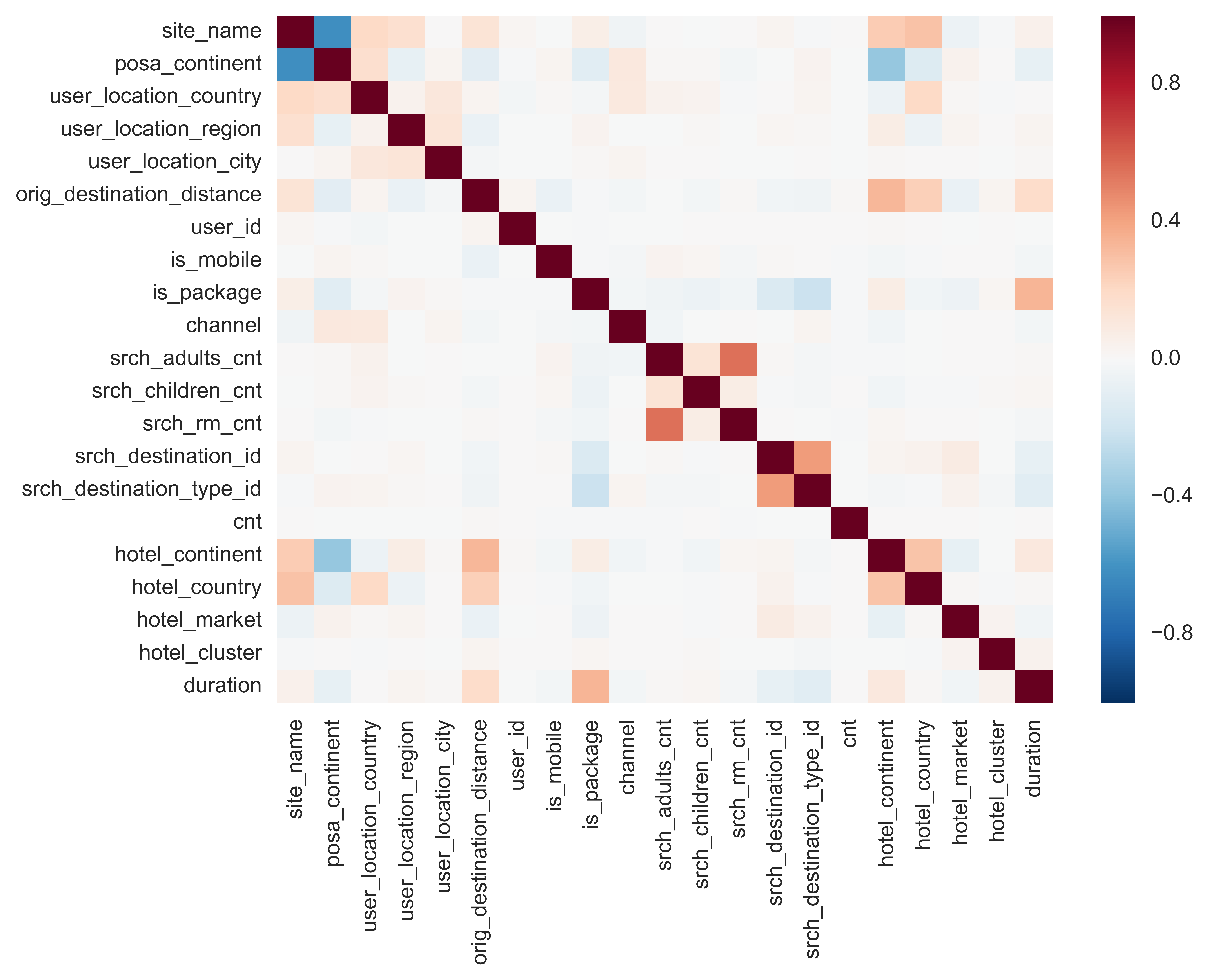}
\end{center}
   \caption{Correlation matrix of features}
\label{fig:cor}
\end{figure}

\begin{itemize}
\item{$hotel\_cluster$ does not seem to have a strong (positive or negative) correlation with any other feature. Thus, methods which model linear relationship between features might not be very successful.}
\item{$orig\_destination\_distance$ has a positive correlation with $duration$ (constructed using $srch\_ci$ and $srch\_co$), which means people who are planning for a long trip tend to go far away from the place of origin.}
\item{$hotel\_continent$ and $posa\_continent$ (which is from where the booking is done) are negatively correlated. This means that people tend to go to continents different from theirs for vacations.}
\item{$duration$ seems to have a strong positive correlation with $is\_package$. This means that people who tend to book hotel for longer duration usually choose hotels as a part of a package.}
\item{$srch\_destination\_id$ has a strong correlation with $srch\_destination\_type\_id$. This is expected as each destination would have an associated type; for example, vacation spot, city, etc.}
\item{$duration$ is also positively correlated with $hotel\_continent$ which means certain continents are preferred for longer duration of stay}
\item{$srch\_rm\_cnt$ has a very strong correlation with $srch\_adults\_cnt$, and to an extent, with $srch\_children\_cnt$ also. This is expected as people tend to take rooms based on how many families/couples are there.}
\end{itemize}

\section{Predictive Task} \label{section:3}
As stated earlier, the goal is to build a personalized recommendation system for the Expedia users based on their preferences and search query. We recommend the top five hotel clusters to the user that he/she is more likely to book than the other ninety-five clusters.

\subsection{The Outcome Variable} \label{section:3.1}
\begin{figure}[h!]
\begin{center}
\includegraphics[width=\linewidth]{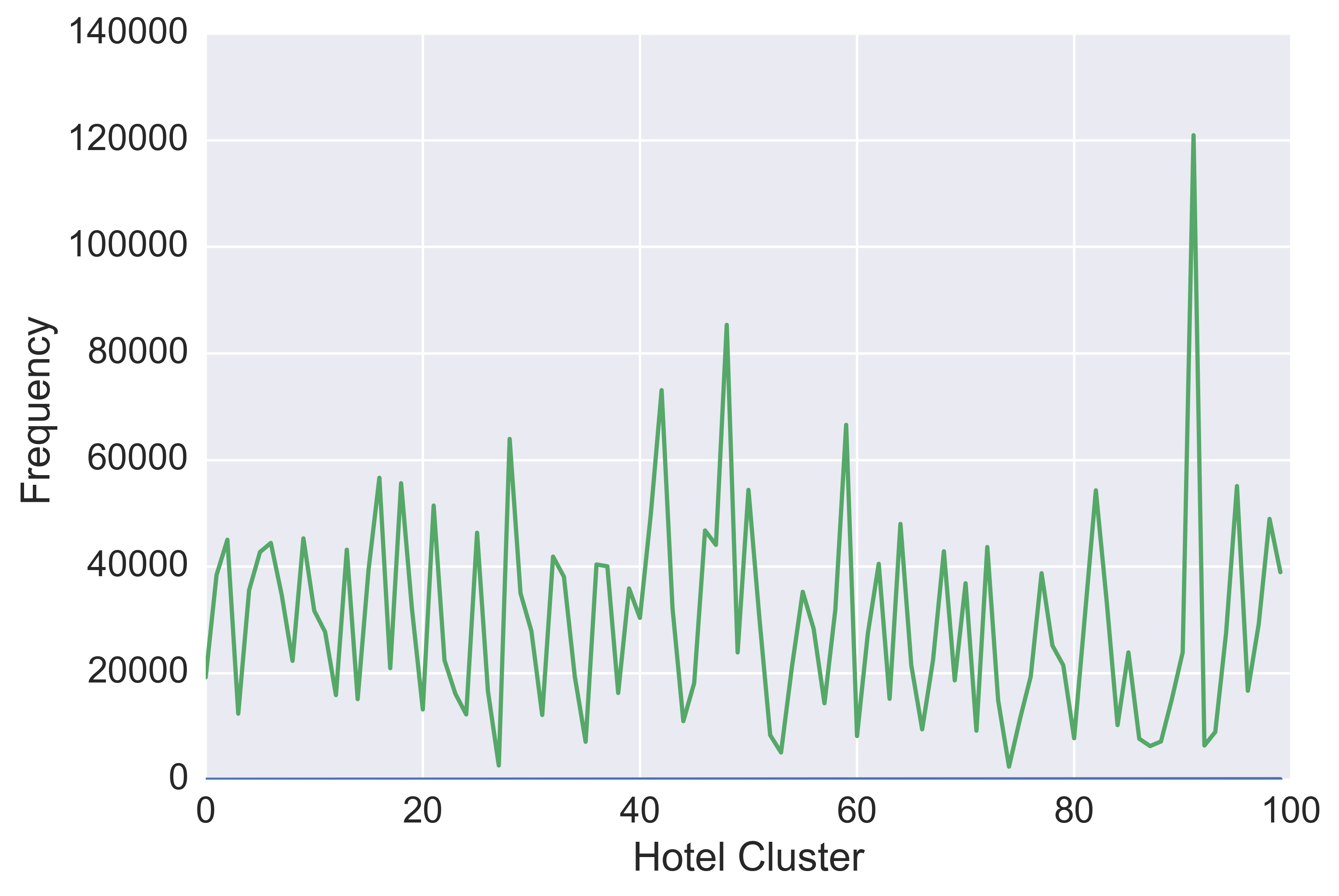}
\end{center}
   \caption{Hotel Cluster Frequency}
\label{fig:hotel_cluster_frequency}
\end{figure}

The outcome of this process is the hotel cluster number - $hotel\_cluster$, which can take values from 0 to 99. Hotel cluster number is a grouping of similar hotels for a search by Expedia. For each observation in the test set, we are expected to recommend the top five hotel clusters for that user-destination-stay combination in a ranked order. As seen in the figure \ref{fig:hotel_cluster_frequency}, the cluster-wise distribution is not even with certain clusters having more instances in the training data than others. 

\subsection{Evaluation Metric} \label{section:3.2}
Given that we are supposed to provide five hotel cluster recommendations, traditional measures like Mean Absolute Error (MAE) and Mean Squared Error (MSE) are not sufficient. Here, we evaluate our predictions on the basis of Mean Average Prediction @ 5 or MAP@5, which can be written as:
$$
MAP@5 = \frac{1}{|U|} \sum\limits_{u=1}^{|U|} \sum\limits_{k=1}^{min(5,n)} P(k)
$$
where $|U|$ is the number of user events, $P(k)$ is the precision at cutoff $k$ and $n$ is the number of predicted hotel clusters.

\subsection{Data Leakage} \label{section:3.3}
During the contest, Expedia confirmed that there was a leak in the data and it had to do with the $orig\_destination\_distance$ attribute. The estimate was that this directly affects approximately 1/3rd of the test data. The contest continued after this without any changes. In other words, Expedia confirmed that one can find $hotel\_clusters$ for the affected rows by matching rows from the train dataset based on the following columns: $user\_location\_country$, $orig\_destination\_distance$, $user\_location\_region$, $user\_location\_city$ and $hotel\_market$. However, predictions based on this will not be 100\% accurate because hotels can change cluster assignments, as hotel popularity and prices have seasonal characteristics.

\section{Learning Models and Methodology} \label{section:4}
In this section, we describe various models that we have used. As described in section \ref{section:2.2}, hotel clusters do not seem to have strong linear correlation with any of the features. Thus, applying techniques that rely on modeling the linear relationship between features would be less fruitful. Hence, we decided to work mostly on tree based methods which are capable of capturing non-linear tendencies in data. Another reason for leaning towards tree based models is that most of the data provided was categorical which can be handled by tree based models without needing any conversion to one-hot form. Discussion on the results of each model is provided in section \ref{section:6}.

\subsection{Baseline} \label{section:4.1}
A baseline is a method that uses heuristics, simple summary statistics or randomness to create predictions for a dataset. We can use these predictions to measure the baseline's performance (e.g., accuracy. MAP @ 5 in our case). This metric will then become what we compare other models' performance against. 

We take a trivial predictor that always predicts the same five clusters as our baseline. These five clusters are the ones that have the maximum occurrence in training data as observed from figure \ref{fig:hotel_cluster_frequency}. Thus, we predict the most frequent hotel clusters provided to us for each user without considering any user preference whatsoever and try to beat this measure by a significant amount. 

\subsection{Random Forest} \label{section:4.2}
A random forest classifier is a meta-estimator that fits a number of decision tree classifiers on various sub-samples of the dataset and use averaging to improve the predictive accuracy and control over-fitting. In Python's sklearn implementation\footnote{http://scikit-learn.org}, the sub-sample size is always the same as the original input sample size but the samples are drawn with replacement by default. As in boosting, many weak learner trees are grown and aggregated into a unified model.

Random forest uses regression trees as building blocks. Using a greedy approach, the feature space is partitioned into a specified number of distinct non-overlapping regions. The number of partitioned regions affects the bias-variance
trade off. Random forests are a bagging method but in them, each grown tree only uses a random sample of $p$ predictors out of all the $n$ predictors. The motivation behind this is to de-correlate trees. When all predictors are
used, all grown trees look like each other. Therefore,
bagged trees are very correlated, which negatively affects
the reduction in variance by averaging. Random forests
solves this problem by considering only $(n−p)/n$ portion
of predictors at each tree split. For $p = n$, random forests
turn into simple bagging.

The parameters to be tuned for random forests are (1)
the number of random predictors, (2) the maximum
number of leaf nodes (which determines how deep a tree
is grown) and (3) the number of trees to be grown and
averaged. Grid Search functionality of Python's sklearn was used for this.

\subsection{SGD Classifier} \label{section:4.3}
Stochastic Gradient Descent (SGD) is a simple yet very efficient approach to discriminative learning of linear classifiers under convex loss functions such as (linear) Support Vector Machines and Logistic Regression. SGD has been successfully applied to large-scale and sparse machine learning problems often encountered in text classification and natural language processing. Given that the data is sparse, the classifiers in this module easily scale to problems with more than $10^5$ training examples and more than $10^5$ features.

The main advantages of SGD are - (1) Efficiency and (2) Ease of Implementation. The disadvantages include - (1) Requirement of a large number of hyper-parameters and (2) Sensitivity to feature scaling. The class SGDClassifier in Python's Sklearn implements a plain stochastic gradient descent learning routine which supports different loss functions and penalties for classification. The parameters to be tuned include - (1) Alpha, (2) number of iterations, (3) epsilon and (4) Learning rate.

\subsection{Naive Bayes} \label{section:4.4}
Naive Bayes methods are a set of supervised learning algorithms based on applying Bayes’ theorem with the “naive” assumption of independence between every pair of features. Given a class variable y and a dependent feature vector $x_1$ through $x_n$, Bayes’ theorem states the following relationship:
$$
P(y | x_1, ..., x_n) = \frac{P(y)P(x_1, ..., x_n|y)}{P(x_1, ..., x_n)}
$$
Using the independent naive independence assumption that 
$$
P(x_i|y,x_1,...,x_{i-1},x_i,x_{i+1},...,x_n) = P(x_i|y), 
$$
for all $i$, this relationship is simplified to 
$$
P(y|x_1,...,x_n) = \frac{P(y)\prod_{i=1}^n P(x_i|y)}{P(x_1,...,x_n)}
$$
Since $P(x_1,...,x_n)$ is a constant, we can use the following classification rule:
$$
\hat{y} = arg max_{y} P(y) \prod\limits_{i=1}^n P(x_i|y)
$$
and we can use Maximum A Posteriori estimation to estimate $P(y)$ and $P(x_i | y)$; the former is then the relative frequency of class $y$ in the training set.

For the purpose of this problem at hand, we used Bernoulli Naive Bayes, which implements the naive Bayes training and classification algorithms for data that is distributed according to multivariate Bernoulli distributions; i.e., there may be multiple features but each one is assumed to be a binary-valued (Bernoulli, boolean) variable. The decision rule for Bernoulli naive Bayes is based on
$$
P(x_i|y) = P(i|y)x_i + (1 - P(i|y))(1-x_i)
$$

\subsection{XG-Boost} \label{section:4.5}
XGBoost \footnote{http://xgboost.readthedocs.io/en/latest/} is short for “Extreme Gradient Boosting”, which refers to using ensembles of weak lerners to make a prediction. XGBoost is based on this model. It is used for supervised learning problems, where we use the training data (with multiple features) $x_i$ to predict a target variable $y_i$.

XGBoost and Random forests are similar in that they both use tree ensembles as the model. The primary difference between them is the way in which they are trained. The random forest treats all its trees uniformly, whereas the boosted trees use different levels of learners to classify progressively difficult samples.

We  try to optimize one level of the tree at a time, since picking the best out of all trees would be impractical. We try to split a leaf into two leaves, and the score it gains is given by the following formula:

$Gain = \frac{1}{2}\bigg[\frac{G_L^2}{H_L + \lambda} + \frac{G_R^2}{H_R + \lambda} - \frac{(GL + GR)^2}{H_L + H_R + \lambda}\bigg] - \gamma$

where the first term refers to the score of the first leaf, the second term refers to the score of the second leaf, and the third term refers to the score before the split into two leaves, i.e. the score at the previous level. The fourth term refers to the regularization on the additional leaf.  

\subsection{Ensemble Learning} \label{section:4.6}
Ensemble learning is the process by which multiple models, such as classifiers or experts, are strategically generated and combined to solve a particular computational intelligence problem. Ensemble learning is primarily used to improve the (classification, prediction, function approximation, etc.) performance of a model, or reduce the likelihood of an unfortunate selection of a poor one. 

In statistics and machine learning, ensemble methods use multiple finite learning algorithms to obtain better predictive performance than could be obtained from any of the constituent learning algorithms alone. For the purpose of our hotel recommendation task at hand, we took an ensemble of the previous models discussed so far:
\begin{itemize}
\item{Random Forest}
\item{Naive Bayes}
\item{SGD Classifier}
\item{XG Boost}
\end{itemize}
The hotel cluster-wise probabilities using each of the above mentioned models were recorded on the test dataset and were then averaged. For each user, the five most probable clusters were chosen as the predictions. While each of the above mentioned models did not give significant results, an ensemble of these proved out to be really well in terms of the MAP @ 5 score.

\subsection{Data Leak Solution} \label{section:4.7}
The data leak solution corresponds to a simple
classification technique leveraging a leakage in
the data. The data leakage was initially proposed by a participant and later confirmed by the Expedia representatives on Kaggle. The features causing the leakage are location
of the user, origin-destination distance, hotel market and
search destination ID. For a test set data point, data leak
solution finds a match in the training data which has the same
values as of test point’s leakage features. If a match for all
features is not found, partial matches are searched. The
test point is assigned to the same cluster of the matched
training point (if found). If no match is found for any feature, then the test point is classified into the most popular clusters (as in the baseline solution).

\subsection{Ensemble Learning Plus Data Leak Solution} \label{section:4.8}
In this approach, we combine the ensemble learning solution with the data leakage solution. If a perfect match is found for leakage features, then the result is populated using the training data. If not, we use the solution returned by the ensemble learning model discussed in \ref{section:4.6}.

\subsection{Distance Matrix Completion and XG Boost} \label{section:4.9}
The feature $orig\_destination\_distance$ is supposed to store the physical distance between the user and the destination. We felt that this feature would be important to improve the accuracy of our hotel cluster prediction, as people generally prefer to go to holiday destinations which are quite far away from their homes. But in the given data, this feature value was missing for most of the cases. As such we decided to run a matrix completion algorithm on the same. For mapping the distances, we used the $user\_location\_city$ as key for the user location and the $hotel\_market$ as the key for the hotel locations. Using these two we initialized a sparse matrix using the existing distance data, with rows where distances were greater than 100kms in order to prevent noisy values. We then normalized it using the max value. This gave us an initial matrix with dimensions of roughly $8000 \times 2000$. For the completion algorithm we chose Iterative SVD using alternating least squares approach, owing to it speed while still maintaining accuracy. After convergence, we rescaled the values using the previous maximum.

\section{Relevant Literature} \label{section:5}
In \cite{Liu2013CombinationOD}, the authors work on a similar problem for personalized hotel recommendations for Expedia. This was part of ICDM 2013 challenge and as such they used a different dataset. Their dataset contained features like hotel price, location, ratings, user purchase history, OTA information for searches. Among these they found price and rating to be useful along with the location feature, which is in direct agreement with our own findings. For the ranking they study various approaches including Boosting Trees, Logistic Regression, SVMRank, Random Forests along with Ensembling over these methods. They achieved their highest score, for an overall ranking of 5, using ensembling with z-score normalization.

In \cite{Levi2012}, the authors approach the problem of hotel recommendation using text based features extracted from reviews. This allows them to tackle the cold-start problem. For this they extract hotel reviews and user information from TripAdvisor and Venere. They extract various kinds of tags from each review based on service, room, food in order to conduct community detection. Apart from this they monitor user nationalities and intent to model preferences, opinion detection using wordnet. They then combine these to get a final review score according to which they rank the hotels.

In \cite{weili}, the authors propose a combination of cluster-based model and collaborative filtering to provide a good recommendation model while also giving a way to tackle cold start problem. For collaborative filtering they compute the user-hotel rating matrix and then use pearson correlation to compute similarities between users. They then cluster similar users together. Whenever a query is made, ratings given by similar users are computed to generate recommendations. In order to handle cold-start problem, they use RankBoost algorithm using four features - standard of rooms, hotel services, location, cost-effectiveness.

In \cite{4811660}, the authors find relations between hotels based on the users who booked them, season in which they are booked, prices, etc. These relations are then filtered on the basis of occurrence, using measures like Jaccard coefficient, and Simpson coefficient. One important aspect of this paper is identifying the transition of preference of users from one hotel to another, i.e. if a user changed his/her preference from one hotel to another, all such occurrences are recorded and used to form transition links between hotels. In a nutshell, this paper places a high importance on the changes in preferences of hotels by users, and uses this data to make hotel predictions for users.
\section{Results and Discussion} \label{section:6}

Table \ref{table:t3} summarizes the results that we have achieved by listing the MAP@5 score for each method used for both the test and training datasets. Note that the $MAP@5$ values for test dataset were obtained after submitting the predictions on Kaggle.

\begin{table}[!ht]
\caption{MAP@5 Scores for Training and Test Sets} 
\centering 
\begin{tabular}{c c c} 
\hline\hline 
Model & Training Set & Test Set \\ [0.5ex] 
\hline 
Baseline & 0.142 & 0.069\\
RF & 0.509 & 0.421\\ 
SGD & 0.354 & 0.301\\
NB & 0.330 & 0.298\\
XGB & 0.533 & 0.432\\
DMC + XGB & 0.566 & 0.463\\
EL & 0.578 & 0.450\\
DL & 0.586 & 0.491\\
EL + DL & 0.592 & 0.496\\ [1ex] 
\hline\\ 
\end{tabular}

\begin{flushleft}RF : Random Forest, SGD : Stochastic Gradient Descent, NB : Naive Bayes, XGB : XGBoost, EL : Ensemble Learning, DL : Data Leak, DMC : Distance Matrix Completion
\end{flushleft}
\label{table:t3}
\end{table}

The baseline model was designed to be trivial and hence, the low performance was expected. We see that random forest classifier performs better than most other classifiers. This is because random forests have critical advantages in this challenge given that (1) they can automatically handle missing values, (2) they do not require binary encoding of features
since they can learn that those features are categorical just
by training, and (3) their weak learner trees are equally
likely to consider all variables. This is very important in
this problem because there are a few features which are
dominant over others, as evident from the exploratory analysis conducted. An exhaustive grid search with cross validation was used to prune the hyper-parameters involved. 

SGD and Naive Bayes clearly out-perform the baseline solution. However, they are not good enough to generalize over the entire data and produce less than satisfactory results compared to other methods. This can be attributed to the fact that most features are either one-hot vectors or have missing values, both of which become hindrances in achieving a good solution via these methods. 

XGBoost has a performance comparative to that of Random Forest but slightly better. This is because both these methods work in a similar manner, handling both categorical features and missing values efficiently. XGBoost automatically learns what is the best direction to go when a value is missing. Its learning mechanism gives a slightly better solution than random forests.

The distance matrix completion method gives a decent result given the fact that distance was one of the most important features. The idea of this method was derived from the blog of first place solution\footnote{https://www.kaggle.com/c/expedia-hotel-recommendations/discussion/21607}. XGBoost was applied in conjunction with distance matrix completion technique because XGBoost outperformed all other naive methods.

Ensemble learning combines various classifiers that we have learned and average out their predicted probabilities leading to better learning on the weak learners. This method averages out the biases and reduces the variance. Given that ensemble learning performs better on difficult cases, we have comparatively better performance using this method than most others.

Surprisingly enough, the data leak solution is one of the best solutions that we obtained. This when combined with the ensemble learning solution gives an even better result. However, the main downside of this approach is the fact that this method does not generalizes well and exists only because the data provided by Expedia was flawed. In other words, we can't rely on this method and it is good only to obtain good scores on the Kaggle leader-board.

The state of the art result has a MAP@5 score of $0.60219$ on Kaggle, with the next best score being $0.53218$. According to the first place solution summary, the winner followed the path of distance matrix completion using gradient descent on spherical law of cosines formula\footnote{https://en.wikipedia.org/wiki/Great-circle\_distance}. Convergence for the gradient descent was not quick at all. Using Nesterov momentum and a gradual transition from squared error to absolute error, the process took about $10^{11}$ iterations and $36$ hours on a dual socket E5-2699v3 with ~700GB RAM. Given that this would have been computationally too expensive for our comparatively primitive machines, we decided against taking this path.

\section{Conclusion and Future Work} \label{section:7}
We have successfully implemented a hotel recommendation system using Expedia's dataset even though most of the data was anonymized which restricted the amount of feature engineering we could do. We ranked the problem at hand as a multi-class classification problem and maximized the probabilities of each cluster, picking the top five clusters at the end.

The most important and challenging part of implementing the solutions was to create and extract meaningful features out of the 38 million data points provided to us. The exploration of data took a long time given the size of data and it helped us extract features that seemed to have high impact on predicting the hotel clusters.

After applying multiple models and techniques, we arrived at the conclusion that Ensemble Learning with Data Leak model performs best giving a score of $0.496$ on test data. This again reaffirms the fact that combining several weak learners has a synergistic affect. Most of our methods involved ranking of clusters by their predicted class probabilities which seems fair. 

Due to the volume of data, we could not replicate the distance matrix completion techniques employed by the first rank solution. This leaves a room for future improvement, wherein we can try replicating the existing code in conjunction with our features, data leak solution and ensemble learning to achieve an even better result. Another direction for future work is to try different ranking methods like RankSVM and RankBoost which perform well but require a lot of resources.

{\small
\bibliographystyle{ieee}
\bibliography{egbib}
}

\end{document}